\documentclass{acmsiggraph}
\usepackage{amsfonts}
\usepackage{bbm}
\usepackage{booktabs}

\title{3DFS: Deformable Dense Depth Fusion and Segmentation for Object Reconstruction from a Handheld Camera}

\author{Tanmay Gupta, Daeyun Shin, Naren Sivagnanadasan, Derek Hoiem \\ University of Illinois Urbana Champaign}
\pdfauthor{***}

\TOGonlineid{0533}


\copyrightyear{2016}

\begin{document}


 \teaser{
   \includegraphics[width=0.9\linewidth]{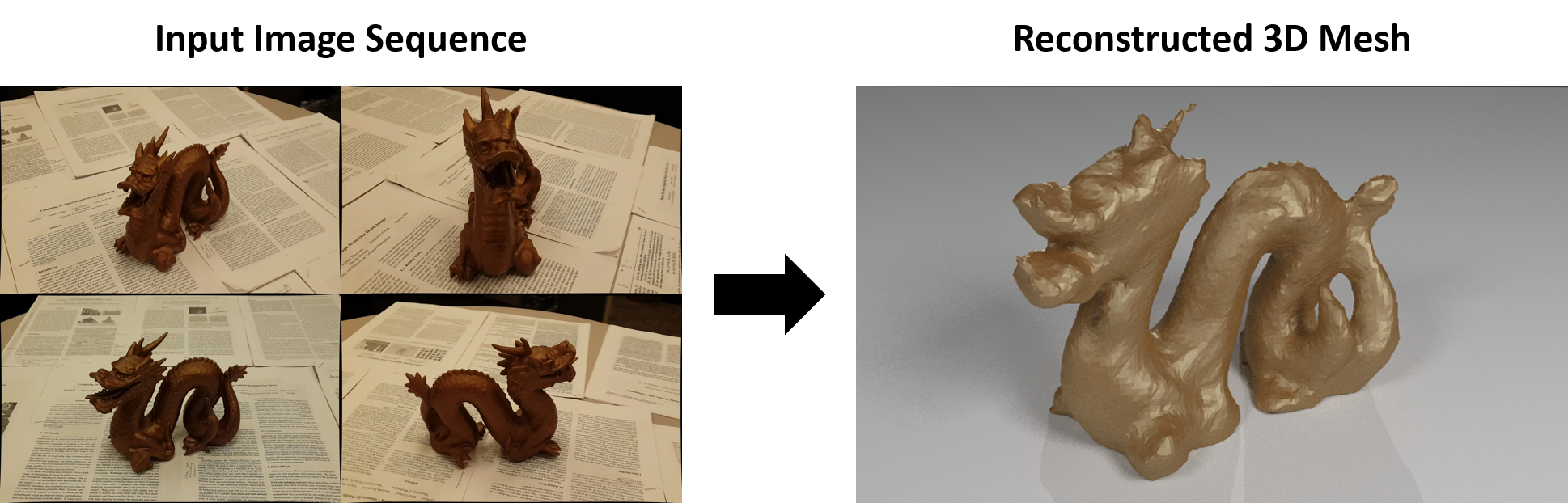}
   \vspace{-0.1in}
   \caption{Our system takes an RGB image sequence and automatically produces a 3D mesh for the object. }
   \vspace{-0.1in}
 }

\maketitle

\begin{abstract}
We propose an approach for 3D reconstruction and segmentation of a single object placed on a flat surface from an input video.  Our approach is to perform dense depth map estimation for multiple views using a proposed objective function that preserves detail.  The resulting depth maps are then fused using a proposed implicit surface function that is robust to estimation error, producing a smooth surface reconstruction of the entire scene.  Finally, the object is segmented from the remaining scene using a proposed 2D-3D segmentation that incorporates image and depth cues with priors and regularization over the 3D volume and 2D segmentations.  We evaluate 3D reconstructions qualitatively on our Object-Videos dataset, comparing to fusion, multiview stereo, and segmentation baselines.  We also quantitatively evaluate the dense depth estimation using the RGBD Scenes V2 dataset~\cite{henry2013patch} and the segmentation using keyframe annotations of the Object-Videos dataset.


\end{abstract}

%
%
%


%
%


\keywordlist

\conceptlist


\section{Introduction}

Our aim is to create a 3D model of a single object recorded by a handheld mobile phone camera.  We assume only that the object is placed on a flat surface and that the object is approximately centered. The ability to easily and accurately create 3D models from handheld cameras has broad applications including virtual reality and 3D printing.  But existing methods are unable to consistently produce 3D models of objects with specularities and irregular shapes without user interaction or controlled capture settings.

3D reconstruction of scenes from image sequences is a well studied problem in computer vision. The underlying idea behind most existing work is that multiple pixel measurements of a point in the scene can be used to triangulate its 3D position. Therefore, reconstruction accuracy hinges on the ability to track a pixel with sub-pixel accuracy. Tracking and matching algorithms, however, assume brightness constancy which breaks down for specular surfaces. Tracking algorithms also falter in textureless regions where neighboring pixels have similar intensities.

MonoFusion \cite{pradeep2013monofusion} and MobileFusion \cite{ondruska2015mobilefusion} demonstrate 3D reconstruction by explicitly modeling surfaces as depth maps and then performing volumetric fusion. However, to achieve real-time performance such methods compromise on accuracy of the depth maps by relying on stereo matching between the live frame and the last key frame. Similarly, volumetric fusion is performed by a weighted average of implicit surface representation of the depth surfaces, such as Truncated Signed Distance Fields (TSDF), computed in different views. Success of such volumetric fusion approaches depends on the accuracy of depth maps. While these techniques work well for depth maps acquired using active sensors like Kinect, they are not robust to localized but often large errors in depth maps that are common with multi-view stereo techniques.

Another important aspect of single object 3D reconstructions is segmentation of the object from the scene. To be useful for 3D printing, the 3D volume occupied by the object needs to be identified. While TSDF gives an estimate of empty regions in space, the cues that distinguish the object from other surfaces come from scene priors and the images. Some approaches iteratively solve for 2D and 3D segmentations~\cite{vogiatzis2005multi} or jointly segment pixels and sparse point clouds~\cite{xiao2007joint}, but none, to the best of our knowledge, perform joint inference over pixels and a dense grid of voxels, which we find to be important for obtaining accurate 3D models.

In this work, we propose to improve the accuracy and robustness of video based multi-view single object 3D reconstruction and segmentation systems by improving surface modeling, volumetric fusion, and segmentation. Our system computes accurate depth maps by posing dense per-pixel depth estimation as an optimization problem which incorporates multi-view stereo cues, sparse point cloud reconstruction from a VSLAM system, and a surface smoothness prior in the form of a rotation invariant bending energy. For 3D reconstruction, we reformulate volumetric fusion of depth maps by getting rid of the \textit{truncation} in TSDF and using a \textit{soft-max} based signed distance function (SDF). Our fusion approach has the benefit of being more robust to errors in depth maps and also produces smoother surfaces. This technique however leads to a shift in the \textit{zero-crossing} of the function field, which we remedy by introducing a novel volume field deformation using a sparse surface point cloud such as that provided by patch-based multi-view stereo~\cite{furukawa2010accurate}. For segmentation, we perform joint inference over pixel and voxel labels in a graph cut optimization framework.  Pixels and voxels impose complementary constraints on the reconstruction. Pixels model the object color which helps distinguish the object surface from other background surfaces and provide cues to surface discontinuities through contour edges. Voxels impose constraints on empty regions, enforce continuity of objects in 3D space and incorporate our scene prior. In addition, surface voxels enforce consistency of pixel labeling across multiple images. We evaluate each of these components to demonstrate good performance even in the presence of specularities and textureless surfaces.

To summarize our contributions:
\begin{enumerate}
\item[1.] We propose a fully automated approach to produce a 3D mesh of a single object placed on a flat surface.
\item[2.] We propose a method to robustly estimate depth surfaces from multi-view stereo cues and a sparse point cloud (optional) which is regularized by a rotation invariant second order bending energy. We demonstrate its performance on our Object-Videos dataset as well as RGBD Scenes v2 dataset and compare against other forms of regularization.
\item[3.] We reformulate volumetric fusion by using a \textit{soft-max} instead of the truncation and weighted averaging for computing the TSDF proposed in \cite{newcombe2011kinectfusion}. To correct the bias in \textit{zero-crossing} we perform a smooth function field deformation using a surface point cloud. Our approach is more robust to errors in depth maps and produces smooth surface meshes.
\item[4.] We formulate joint 2D image segmentation and volumetric 3D reconstruction as the task of assigning discrete labels $\{object,background\}$ to every pixel and $\{object,background,empty\}$ to every voxel $v$.The discrete optimization is solved using graph cut with $\alpha$-expansion procedure under constraints imposed by pixels, voxels and scene priors.
\item[5.] We evaluate our entire system through pixel segmentation accuracy on Object-Videos dataset which is a video dataset of 12 objects recorded by a mobile phone camera. We provide ground-truth segmentation masks for selected frames to encourage advancement of state-of-the-art on this task.
\end{enumerate}

\section{Related Work}

We can group relevant literature into two broad areas: 3D reconstruction and multi-view segmentation.

\noindent
\textbf{3D Reconstruction:} Most of the existing work in this area focuses on reconstruction of an entire scene. Different approaches cater to different reconstruction resolution requirements and size of the scenes. Single object reconstruction mostly falls within the purview of literature that deals with small sized indoor scenes. MonoFusion \cite{pradeep2013monofusion} and MobileFusion \cite{ondruska2015mobilefusion} use stereo matching between the live frame and last selected key-frame to generate depth maps. The final surface reconstruction is computed by extracting zero-level set of the Signed Distance Field (SDF) obtained by weighted averaging of TSDFs for different views. This approach to volumetric fusion, popularized by KinectFusion \cite{newcombe2011kinectfusion}, has demonstrated good performance for fusing accurate depth measurements from active sensors like the Kinect. However, this method has two drawbacks. First, due to weighted averaging, the TSDFs are forced to be asymmetric to avoid changing the zero-level set. This in turn hinders their ability to collect evidence of occupancy for voxels behind the visible surfaces. Second, this approach relies on accurate depth surface estimates which, while generally true for active sensors operating in indoor scenes, does not generalize to stereo based methods in presence of specular reflection and texture-less surfaces. To counter these problems we propose an alternative scheme that relies on maximum signed distance to the visible surfaces. For robustness we use a \textit{soft-max} instead of a hard maximum. This is followed by \textit{zero-crossing} correction using a smooth deformation field generated using a sparse point cloud provided by PMVS \cite{furukawa2010accurate}. Our joint 2D-3D segmentation also reduces dependence on the depth maps by utilizing multiple cues and hence improves the overall robustness of the system.

DTAM \cite{newcombe2011dtam} proposes a novel and robust alternative to pairwise stereo for computing dense per-pixel depth maps. They pose the problem of computing depth at evey pixel as a two step iterative optimization problem, where the first step ensures consistency of depth estimates with photometric evidence integrated over multiple frames (\textit{cost volume}) and the second step provides a first order spatial regularization. Inspired by DTAM, we also pose depth surface computation as a two step iterative optimization problem but with some important improvements and simplifications. The major differences are: (1) relaxation of frontal-planar assumption imposed by the first order regularization; (2) removal of Huber loss on smoothness term and appropriate compensation through spatially varying weights that turns the optimization problem in the second step to linear least squares; (3) replacement of pixel based photometric error by patch based Zero-mean Normalized Cross Correlation (ZNCC) which is more robust to brightness changes, while computing the cost volume.

Another approach to dense 3D reconstruction is to begin with a sparse point cloud, increase density by propagating measurements to nearby points using techniques like PMVS \cite{furukawa2010accurate} and then fit a mesh to this semi-dense reconstruction \cite{kazhdan2006poisson}. However to expand the point cloud, techniques like PMVS perform only local operations that may not produce a globally consistent result. It is also difficult to enforce surface regularization.

Camera pose estimation is an integral component of any of the above mentioned methods. Direct monocular slam approaches like LSD-SLAM \cite{engel2014lsd} and DTAM, which directly minimize photometric error to register live frame with the last selected key-frame  have been shown to be more robust than feature tracking based approaches such as PTAM \cite{klein2009parallel}. LSD-SLAM also provides semi-dense depth estimates with inverse depth variance which expresses belief about the accuracy of the estimates.

\begin{figure*} [t]
\begin{center}
\includegraphics[width=0.9\linewidth]{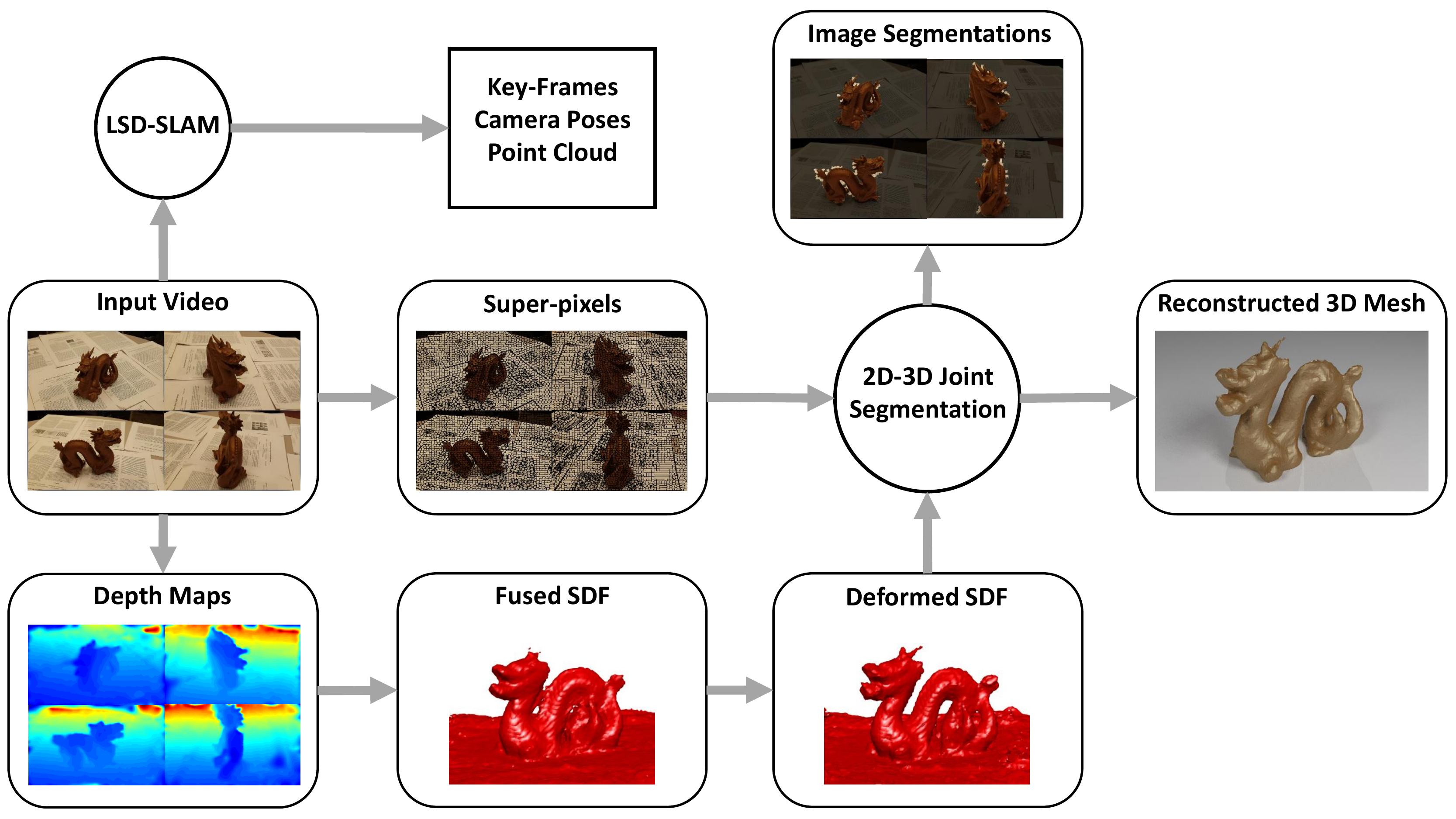}
\end{center}
   \vspace{-0.1in}
   \caption{
   System overview: First, LSD-SLAM is used to estimate camera poses and to select key-frames from the input video. Then we compute dense depth maps which are volumetrically fused using a robust 2 stage process involving computation of a \textit{softmax} based \textit{signed distance function} followed by deformation using PMVS point cloud. Finally, a joint 2D-3D segmentation is performed to obtain a 3D reconstruction with consistent 2D segmentations.}
   \vspace{-0.1in}
\label{fig:overview}
\end{figure*}

\noindent
\textbf{Multi-view Segmentation:} The dominant approach for multi-view segmentation involves finding optimal  labels for some subset of \{\textit{pixels}, \textit{superpixels}, \textit{3D points}, \textit{voxels}\} that minimizes an energy function that encodes task specific priors and evidence from data.

The work most closely related to ours is \cite{vogiatzis2005multi}, where 3D reconstruction is posed as a volumetric Graph Cut \cite{boykov2004experimental} optimization problem defined over a 6-connected grid of voxels. Foreground and background color distributions were modeled from images and the unary term for a voxel was defined in terms of the average posterior probabilities of the pixels that the voxel projected to. The foreground voxels were then projected onto the images to get segmentation masks and the color model was updated. Similarly, \cite{djelouah2013multi} propose an approach for segmenting an object from a video by performing graph cut over superpixels and sparse 3D points. They have explicit edges between 3D points and superpixels to ensure multi-view coherence, and edges between superpixels across frames, which are related by optical flow, to model temporal consistency. Both the approaches generate image segmentation masks, but, while the first method generates a volumetric reconstruction by labeling voxels, the second approach only segments a sparse point cloud. Our formulation borrows the idea of jointly labeling superpixels and a voxel grid with edges between them to enforce consistency, but we incorporate rich information from volumetrically fused dense depth maps computed in early stages of our system and scene priors to get good quality reconstructions of objects with complicated shapes and varying material properties. Our system also differs from these approaches in that we use a richer set of labels that distinguish between the voxels that belong to the object of interest and those belonging to other objects in the scene, namely voxels behind the support surface. This distinction allows a more complete and accurate reasoning about the scene.

Like us, \cite{kowdle2012multiple} attempts to exploit the rich evidence available in dense depth maps, using stereo and color based appearance cues to first compute dense piecewise planar depth maps for every view. Then they assign a foreground or background label to polygons in every image independently. Finally, they fuse these independent segmentations using multi-view reasoning. Each of these steps is performed by $\alpha$ expansion based energy minimization with carefully chosen terms. Unlike us, they only generate image segmentations. Also, the piecewise planar assumption also does not hold for many objects of interest.

An alternative to graph cut optimization was proposed in \cite{xiao2007joint}, where joint segmentation is performed over 2D pixels and 3D scene points using affinity propagation. A graph is constructed over \textit{joint} nodes each of which comprises of a 3D point and its image projections in different views. Similarity between nodes is defined using 3D features like spatial proximity and angle between normal directions, and 2D features like color differences and Kullback-Leibler divergence between patch histograms. This method, however, requires user initialization and includes only sparse 3D points in the optimization. 

All the above approaches perform 3D segmentation by converting the problem to a discrete optimization. \cite{kolev2009continuous} formulate volumetric reconstruction as that of minimizing a continuous convex functional by relaxing binary labels to lie in $[0,1]$, and they show quantitative improvement over discrete graph cut based approaches.

\section{Approach}

Our system takes as input an image sequence and automatically generates a volumetric reconstruction of the object, depth surfaces, and segmentation masks for selected key-frames, as shown in Fig.~\ref{fig:overview}. There are four main stages:

\noindent
\textbf{Pose Estimation:} A state-of-the-art VSLAM system LSD-SLAM \cite{engel2014lsd} is used to get camera poses, key-frame selection, and semi-dense depth maps which are used in later stages of the system.

\noindent
\textbf{Surface Modeling:} The visible surfaces from each view are modeled as dense depth maps. The procedure, inspired by DTAM \cite{newcombe2011dtam}, involves minimizing an objective function comprising of a \textit{cost volume} based data term regularized by a spatially varying linearized bending energy. We experiment with second order bending energies and compare to the first order regularization used in \cite{newcombe2011dtam}.

\noindent
\textbf{Volumetric Fusion:} Depth maps from the previous stage are fused together volumetrically to generate a signed function field over a voxel grid. The function field indicates normalized and clipped signed distance of every voxel to its nearest surface. The fusion, however, introduces a bias in the function field and shifts the zero-crossing. PMVS \cite{furukawa2010accurate} points which are known to lie on the surface are used to correct the bias by deforming the function field where the deformation is modeled by a radial basis expansion.

\noindent
\textbf{Joint 2D-3D Segmentation:}
A joint segmentation of all key-frame images and a common voxel map is performed using graph cuts with $\alpha$-expansion over the set of all pixel and voxel nodes. The  pixel unary models the color of the object and background regions. The voxel unary enforces constraints on the empty regions (using the SDF) and incorporates scene prior by encouraging voxels below the fitted plane to belong to the background. Pixel-pixel and voxel-voxel pairwise terms are used to impose smoothness in 2D (edge-aware) and 3D space respectively. Finally, edges connecting pixel nodes to sparse surface voxel nodes (identified by hashing sparse 3D points generated by LSD-SLAM into the voxel grid) implicitly enforce consistent labeling across views.

\subsection{Pose Estimation}
Camera poses and an initial sparse point cloud with visibility information for each point are required in later stages. Camera focal length, optical center, and radial distortion parameters are precomputed using a checkerboard pattern. The video frames are also corrected for radial distortion. LSD-SLAM \cite{engel2014lsd} with loop closure is then used to select a set of key-frames, compute camera poses for every frame, and generate semi-dense depth maps with corresponding estimates of inverse depth variance for all key-frames.

\begin{figure*} [t]
\begin{center}
\includegraphics[width=0.9\linewidth]{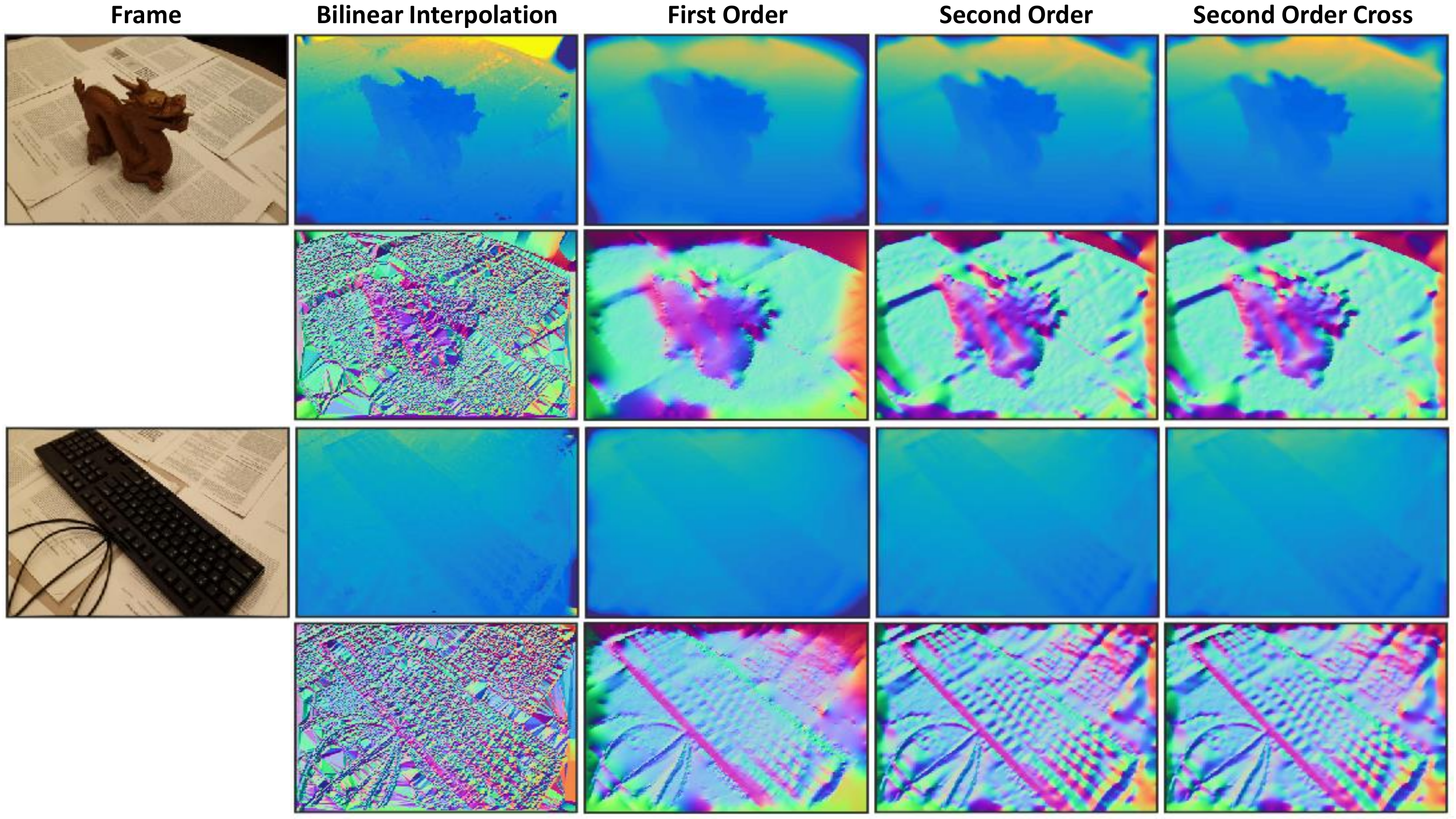}
\end{center}
    \vspace{-0.1in}
   \caption{Visualization of depth maps and surface normals generated for videos in the Object-Videos Dataset using our depth surface reconstruction method. Smoothness constraints in our optimization make the surface normals well behaved. Higher order bending energy helps preserve more fine details such as those on the dragon's head and keyboard's keys and wire.}
   \vspace{-0.1in}
\label{fig:depthNormal}
\end{figure*}

\subsection{Surface Modeling} \label{denseDepth}
Let the set of all pixels in the current reference image $I_r$ be denoted by $\mathcal{P}$. The pixel coordinates will be referred to as $(u,v)$ or by a 2 dimensional vector $\textbf{u}$. The problem of estimating depth surface reduces to assigning a depth value $\xi(\textbf{u})$ to each pixel $\textbf{u} \in \mathcal{P}$ such that the assignment results in low photometric error, agrees with the depth measurements $\hat{\xi}(\textbf{u})$ wherever available and is mostly smooth barring depth discontinuities. In our system these depth measurements are provided by LSD-SLAM for a set of pixels $\mathcal{T}$. LSD-SLAM also provides an estimate of the inverse depth variance $\sigma(\textbf{u})$ for all $\textbf{u}\in \mathcal{T}$.

\begin{figure*} [t]
\begin{center}
\includegraphics[width=0.9\linewidth]{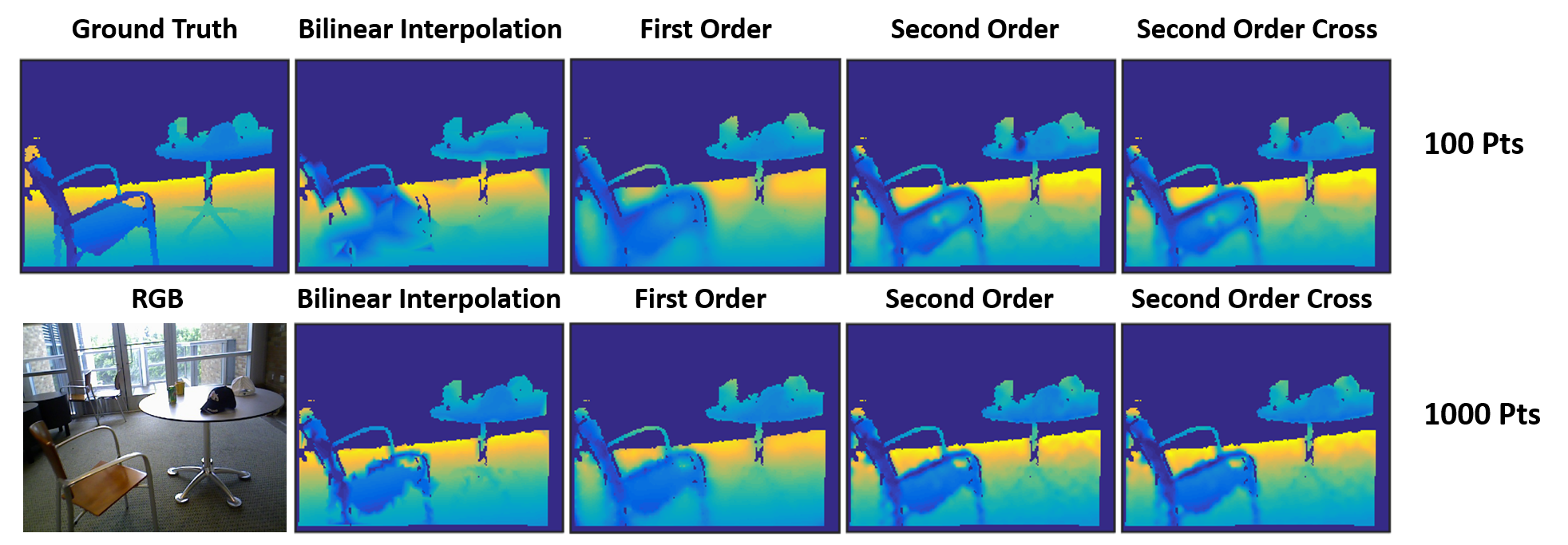}
\end{center}
    \vspace{-0.1in}
   \caption{Effect of different smoothness terms and number of measurement points available on depth map estimation for RGBD Scenes V2 dataset. Our method is robust in presence of specularity and textureless surfaces. The second order bending energies allow the reconstruction of long thin surfaces like the arm rest on the chair.}
    \vspace{-0.1in}
\label{fig:rgbd_qual}
\end{figure*}

Inspired by DTAM \cite{newcombe2011dtam}, we begin by constructing the cost volume $C(\textbf{u},d)$ which specifies the average photometric error of pixel $\textbf{u}$ with respect to neighboring frames for some discrete depth candidates $d\in \mathcal{D}$. Instead of simply using the difference between the pixel intensities, we use a measure based on Zero-mean Normalized Cross-Correlation (ZNCC) between patches of size $3\times 3$. ZNCC has two advantages: (i) removal of brightness constancy assumption; and (ii) robustness to false matches between pixels with similar intensities but different neighborhoods. For a selected set of neighboring frames $N(m)$, the cost volume is given by
\begin{equation}
C(\textbf{u},\xi(\textbf{u})) = \frac{1}{|N(m)|} \sum_{I\in N(m)} \frac{1-Z(\textbf{u},\xi(\textbf{u});I,I_r)}{2}
\end{equation}
where $Z(u,\xi(\textbf{u});I,I_r)$ is the ZNCC between a patch around $\textbf{u}$ in $I_r$ and the patch around its reprojection in $I$ assuming depth $\xi(\textbf{u})$. Depth estimation is now formulated as the following optimization problem
\begin{multline} \label{eq:optProblem}
\min_{\xi}\sum_{\textbf{u}\in \mathcal{P}} C(\textbf{u},\xi(\textbf{u})) +\\
\sum_{\textbf{u}\in \mathcal{T}}\lambda_m(\textbf{u})(\xi(\textbf{u})-\hat{\xi}(\textbf{u}))^2
+\sum_{\textbf{u}\in \mathcal{P}} \lambda_s(\textbf{u})\mathcal{B}(\textbf{u})
\end{multline}
where $\mathcal{B}(\textbf{u})$ is the surface bending energy at $\textbf{u}$ and $\lambda_s(\textbf{u})$ is a spatially varying weight that prevents smoothing across edges. For ease of optimization, we restrict the choice of bending energy to those which can be approximated by finite differences. Ideally, bending energy must also be invariant to in-plane rotation of the camera. One such bending energy has the form
\begin{equation}\label{eq:bendingEnergy}
\mathcal{B}(\textbf{u}) = \left(\frac{\partial^2 \xi(\textbf{u})}{\partial u^2}\right)^2 + \left(\frac{\partial^2 \xi(\textbf{u})}{\partial v^2}\right)^2 + 2\left(\frac{\partial^2 \xi(\textbf{u})}{\partial u \partial v}\right)^2
\end{equation}

Experiments show that dropping the cross term produces similar results but considerably speeds up the optimization procedure. A commonly used substitute, which is also used in \cite{newcombe2011dtam}, is squared norm of the gradient $\|\nabla\xi(\textbf{u})\|^2$. However, this imposes an often misleading bias towards frontal-planar surfaces.

The optimization problem in Equation \ref{eq:optProblem} is solved using a two-step iterative procedure, similar to \cite{newcombe2011dtam}, by introducing auxiliary variables $\alpha$ and a coupling term whose role in the optimization is controlled by parameter $\theta$
\begin{multline}
\min_{\xi,\alpha}\sum_{\textbf{u}\in \mathcal{P}} \lambda_c C(\textbf{u},\alpha(\textbf{u})) + \frac{1}{2\theta}(\alpha(\textbf{u})-\xi(\textbf{u}))^2 +\\
\sum_{\textbf{u}\in \mathcal{T}}\lambda_m(\textbf{u})(\xi(\textbf{u})-\hat{\xi}(\textbf{u}))^2
+\sum_{\textbf{u}\in \mathcal{P}} \lambda_s(\textbf{u})\mathcal{B}(\textbf{u})
\end{multline}
where $\lambda_c$ is a constant that is empirically determined and $\lambda_m(\textbf{u})$ is a spatially varying weight that depends on the reliability of the measurement at $\textbf{u}$.

\noindent
The first step involves minimizing the data term by point-wise search; while the second step solves for the surface that is consistent with the estimates in the first step, agrees with the measurement, and has low bending energy. 

\noindent
\textbf{Step 1} involves a per-pixel point-wise search over the discrete depth candidates to obtain photo-consistent depth estimates
\begin{align}\label{step1}
\hat{\alpha}(\textbf{u}) &= \arg\min_{d\in \mathcal{D}} E_{aux}^{(t)}(\textbf{u},d,\xi^{(t-1)})
\end{align}
\begin{align}
E_{aux}^{(t)}(\textbf{u},d,\xi^{(t-1)}) &= \lambda_c C(\textbf{u},d)
+  \frac{1}{2\theta^{(t)}}(d-\xi^{(t-1)}(\textbf{u}))^2
\end{align}

where $\theta^{(t)}$ is decreased after every iteration to increase the coupling between the discrete auxiliary variables and continuous depth estimates. This step is identical to \cite{newcombe2011dtam} except that our final estimate in the current iteration $\alpha^{(t)}$ is obtained by median filtering the depth image $\hat{\alpha}$ to increase robustness to outliers and then performing a single Newton step using the numerical gradient of $E_{aux}^{(t)}$
\begin{align}
\alpha^{(t)}(\textbf{u}) \leftarrow \alpha^{(t)}(\textbf{u}) - \eta \frac{\nabla E_{aux}^{(t)}(\textbf{u},\alpha^{(t)},\xi^{(t-1)})}{\nabla^2 E_{aux}^{(t)}(\textbf{u},\alpha^{(t)},\xi^{(t-1)})}
\end{align}
where $\eta$ is the learning rate. We use $\eta=0.01$ for our experiments. $\lambda_c=10$ was used for RGBD Scenes dataset and $\lambda_c=1000$ was used for Object-Videos dataset.

\noindent
\textbf{Step 2} enforces smoothness and consistency with depth measurements while being tied to photo-consistent depth estimates obtained in the previous step.
\begin{align}
\begin{split}
\xi^{(t)} = \arg\min_{\xi}\sum_{\textbf{u}\in \mathcal{T}}\lambda_m(\textbf{u})(\xi(\textbf{u})-\hat{\xi}(\textbf{u}))^2 + \\
\sum_{\textbf{u}\in \mathcal{P}} \frac{\lambda_a(\textbf{u})}{2\theta^{(t)}}(\alpha^{(t)}(\textbf{u})-\xi(\textbf{u}))^2 + \lambda_s^{(t)}(\textbf{u})\mathcal{B}(\textbf{u})
\end{split}
\end{align}
In contrast to \cite{newcombe2011dtam}, we remove the Huber loss on the smoothness term which turns our optimization problem into linear least squares thereby improving efficiency. We deal with outliers and large depth discontinuities by employing two techniques: 1) median filtering the optimal solution to equation (\ref{step1}); 2) appropriate choice of spatially varying constants $\lambda_m, \lambda_a$ and $\lambda_s$:

\begin{enumerate}
\item[i.] \textit{\textbf{Measurement weights}}: $\lambda_m(\textbf{u})$ is chosen to be inversely proportional to $\sigma(\textbf{u})$, the inverse-depth variance estimate for each measurement provided by LSD-SLAM. We use $0.1$ as the constant of proportionality.

\item[ii.] \textit{\textbf{Smoothness weights}}: Since we remove the Huber loss on smoothness term, we gradually increase $\lambda_s^t(u)$ with every iteration so that by the time the smoothness term dominates, $\xi$ is already very close to the true solution. The weight also depends upon edge strength to make smoothing edge aware. Specifically we use
\begin{equation}
\lambda_s^{(t)}(\textbf{u}) = \gamma^{t-1} \lambda_s^{(0)}e^{-c_{\text{edge}}\; Edge(\textbf{u})}
\end{equation}
where $Edge(\textbf{u})$ is the edge strength obtained from Structured Edge Detector ~\cite{dollar2013structured} which was trained to detect object contours in images. For our experiments we used $\lambda_s^{(0)}=0.01$ and $c_{\text{edge}}=5$. $\gamma=1.6$ was used for RGBD Scenes dataset, whereas $\gamma=1.8$ was used for Object-Videos dataset.
\item[iii.] \textit{\textbf{Coupling weights}}: Similar to \cite{newcombe2011dtam}, we decrease $\theta$ with every iteration to increase coupling
\begin{equation}
\theta^{(t)} = (1-\beta t)\theta^{(t-1)}
\end{equation}
In addition, photometric error based spatially varying weights are used to propagate depth estimates from more \textit{confident} regions to less \textit{confident} ones such as textureless and specular surfaces.
\begin{equation}
\lambda_a(\textbf{u}) = e^{-c_{\text{conf}}\; \rho(\textbf{u})}
\end{equation}
where $\rho(\textbf{u})$ is the minimum photometric error stored in the cost volume at pixel $\textbf{u}$
\begin{equation}
\rho(\textbf{u}) = \min_{d}C(\textbf{u},d)
\end{equation}
In our experiments, we use $\theta^{(0)}=1$, $\beta=0.03$, and $c_{\text{conf}}=5$.
\end{enumerate}

Note that with the exception of $\lambda_c$ and $\gamma$, the same set of parameters generalized to two datasets with very different scales.

\subsection{Volumetric Fusion}
The depth maps estimated for the key-frames are fused together to get a volumetric reconstruction of the scene represented by an implicit surface function field $S(\textbf{v})$ defined on a voxel grid, $\mathcal{V}$. A desired property of such a function field is that $S(\textbf{v})$ takes a positive value in empty regions and negative value otherwise. The surface mesh can then be extracted as the zero-level set of $S$. Let $S_i(\textbf{v})$ be the Signed Distance Function (SDF) computed over $\mathcal{V}$ due to $i^{th}$ key-frame. Given $K$ key-frames, the actual choice of the functional mapping from $\{S_1(\textbf{v}),\cdots,S_K(\textbf{v})\}$ to $S(\textbf{v})$ is closely related to the choice of the mapping from the $i^{th}$ key-frame's depth map $\xi_i$ to $S_i(\textbf{v})$. Another important requirement from $S(\textbf{v})$ is that its zero-crossing lies on the object's surface. If the depth maps were very accurate, then for any choice of $S_i(\textbf{v})$ which assigns a negative value to voxels behind the depth surface and positive otherwise, $S(\textbf{v})$ can simply be computed as follows
\begin{equation}
S(\textbf{v}) = \max_{i\in \{1,\cdots,K\}} S_i(\textbf{v})
\end{equation}
The surface extracted from $S(\textbf{v})$ will be guaranteed to be accurate up to the resolution of the voxel grid. Sub-voxel accuracy however still depends on the computation of $S_i$. Naturally smooth and more accurate surfaces can be expected if $S_i(\textbf{v})$ is proportional to the distance of $\textbf{v}$ from the closest depth surface.

\begin{figure} [t]
\begin{center}
\includegraphics[width=0.9\linewidth]{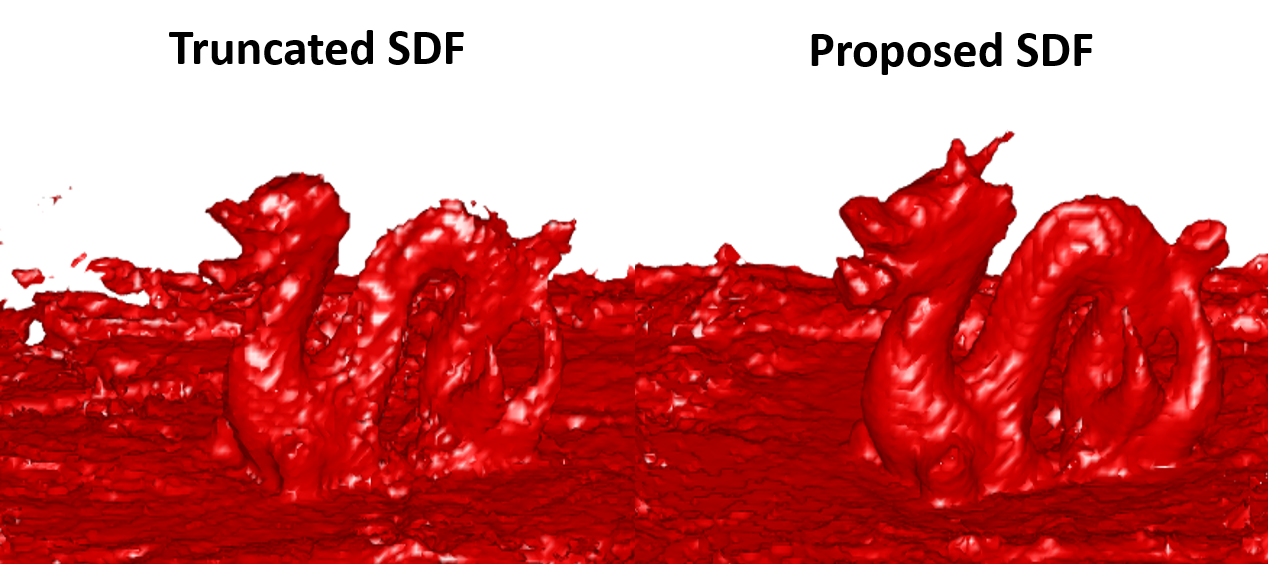}
\end{center}
\vspace{-0.1in}
   \caption{Comparison of TSDF with the proposed robust fusion scheme involving deformation of Softmax SDF. The proposed SDF is more robust to errors in depth estimation and produces smoother more accurate surfaces.}
   \vspace{-0.1in}
\label{fig:tsdfVSssdf}
\end{figure}

\begin{figure} [t]
\begin{center}
\includegraphics[width=0.9\linewidth]{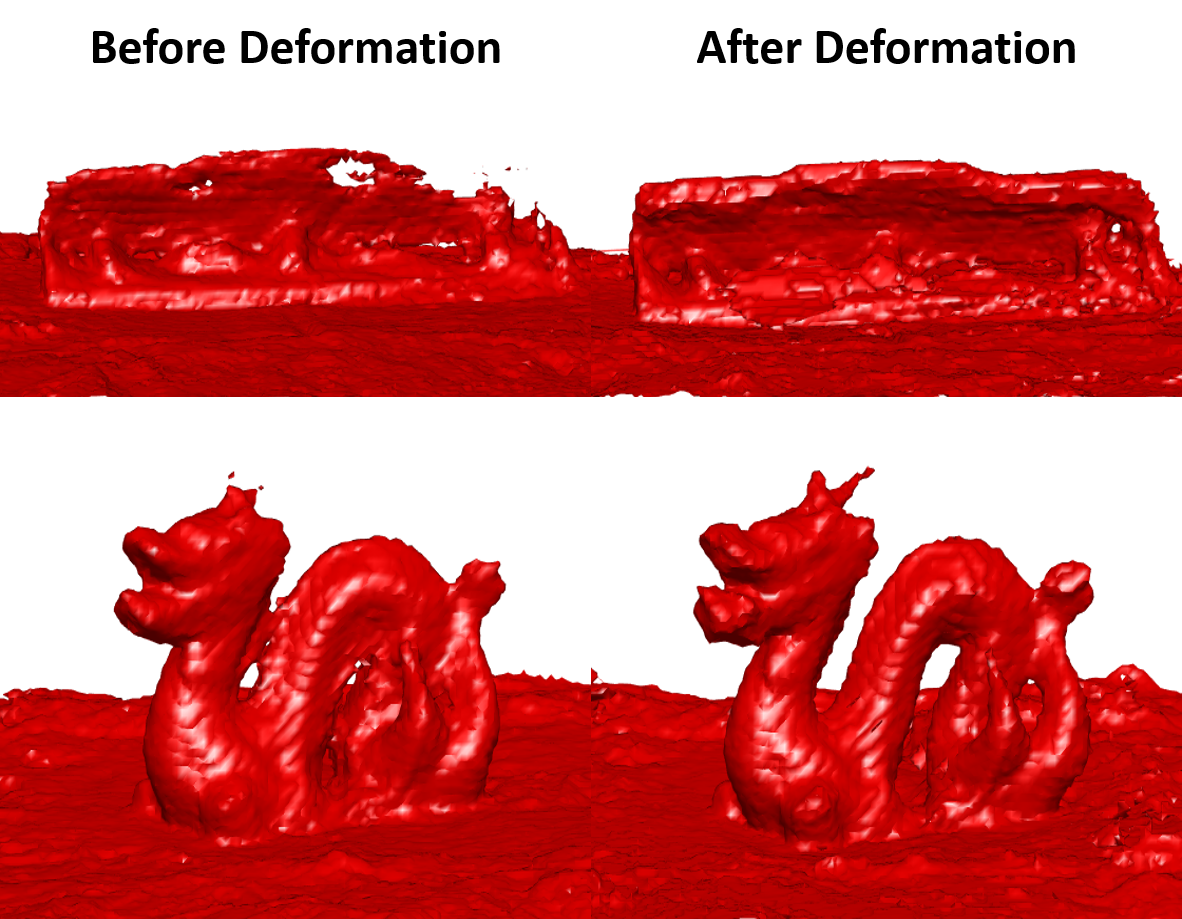}
\end{center}
\vspace{-0.1in}
   \caption{Visualization of the zero-level set of the fused SDF before and after deformation. Notice how the zero-crossing correction recovered the thin structures like the lever arm of the paper punch and horns of the dragon}
   \vspace{-0.1in}
\label{fig:deformation}
\end{figure}

However, since the estimated depth maps often have local errors in regions with textureless surfaces or specularities, the above computations need to be suitably modified for robustness. We propose to decompose volumetric fusion into two parts: i) robust fusion, and ii) zero-crossing or bias correction. The first step produces SDFs that are robust to errors in depth map or camera pose estimation at the cost of incurring a bias or a shift in the zero-crossing. The second step corrects for this bias by applying a smooth deformation field to force the function value close to zero at a set of points known to lie on the object surface. We compare our proposed approach to TSDF~\cite{newcombe2011kinectfusion} in Fig.~\ref{fig:tsdfVSssdf}. Fig.~\ref{fig:deformation} shows the effect and importance of zero-crossing correction.

\subsubsection{Robust Fusion}
Let $\textbf{q}_i(\textbf{v})$ be the point where the ray from the $i^{th}$ camera center to $\textbf{v}$ intersects the depth surface due to $\xi_i$. $S_i(\textbf{v})$ is simply given by
\begin{equation}
S_i(\textbf{v}) = \min\left(1, \frac{\|\mathbf{\eta}_i(\textbf{v})\|_2}{\mu}\right) \times
\text{sign}\left(\eta_{iz}(\textbf{v})\right)
\end{equation}
where $\mathbf{\eta}_i(\textbf{v}) = \textbf{T}_i(\textbf{q}_i(\textbf{v})) - \textbf{T}_i(\textbf{v})$ with $\textbf{T}_i$ being the transformation from world coordinates to camera coordinates. Here $\mu$ is a constant used to clip the SDF value further away from the surface.

The SDFs from all key-frames are fused using a soft maximum operation defined as follows
\begin{equation}
S'(\textbf{v}) = \frac{\sum_{i=1}^{K}S_i(\textbf{v})e^{h S_i(\textbf{v})}}{\sum_{i=1}^{K}e^{h S_i(\textbf{v})}}
\end{equation}

where $h$ is a hardness constant. We use $h=10$ for all our experiments. Soft maximum produces level surfaces that are naturally smooth and trades the property of preserving the zero-crossings for robustness to errors in depth and camera parameter estimates. Note that this is in contrast to \cite{newcombe2011kinectfusion} where $S_i$ were truncated to zero beyond a depth of $\mu$ behind the depth surface in order to preserve zero-crossing as it relied on a weighted averaging of $S_i$s. Views in which a voxel was more directly in the line of sight were given a higher weight while fusing the SDFs. Such a weighing is implicit in our soft-max formulation since voxels tend to have higher $S_i(\textbf{v})$ in views where it is directly in the line of sight.  

\subsubsection{Zero-crossing Correction}

To correct for the bias in $S'$, we use a sparse set of 3D surface points, $\mathcal{M}$ generated by PMVS \cite{furukawa2010accurate}. We pose this problem as that of finding a deformation field $\Delta S(\textbf{v})$ such that
\begin{equation}
S(\textbf{p}) = S'(\textbf{p}) + \Delta S(\textbf{p}) = 0 \: \forall \: \textbf{p} \in \mathcal{M}
\end{equation}
We choose to parametrize $\Delta S$ by a radial basis expansion
\begin{equation}
\Delta S(\textbf{v}) = \sum_{j=1}^{C} a_j \phi(\|\textbf{v} - \textbf{c}_j\|_2)
\end{equation}
where $\textbf{c}_j$ is a set of control points, $\textbf{a}$ denotes the coefficients of expansion, and $\phi$ is any radial basis kernel. For our experiments, we use $\phi(x) = e^{-x^2/\sigma^2}$ with $\sigma=0.1$. For a voxel grid of maximum grid dimension $200$, we use $C=4000$ randomly sampled control points out of which half are sampled in the region $\{\textbf{v}\in \mathcal{V} | -0.5 < S'(\textbf{v}) < 0.5\}$ and the other half outside this region. The coefficients are obtained by minimizing the following least squares objective function
\begin{multline}
\sum_{\textbf{p}\in \mathcal{M}'_1}
\left(
S'(\textbf{p}) +
\sum_{j=1}^{C}a_j\phi(\|\textbf{p} - \textbf{c}_j\|_2)
\right)^2 + \\
\sum_{\textbf{p}\in \mathcal{M}'_2}
\left(
\sum_{j=1}^{C}a_j\phi(\|\textbf{p} - \textbf{c}_j\|_2)
\right)^2 +
\lambda\frac{\|\textbf{a}\|^2}{2}
\end{multline}
The first term enforces $S(\textbf{p})=0$ for $\textbf{p}\in\mathcal{M}'_1$ which contains $500$ points randomly sampled from $\mathcal{M}$. The second term constrains the deformation to be zero at points $\textbf{p}\in \mathcal{M}'_2$ which contains 500 points each sampled from $0.9$ and $-0.9$ level sets of $S'$. The third term is used to regularize the deformation.

\subsection{Joint 2D-3D Segmentation}
Given a set of key-frame images $\mathcal{I}$, their camera pose estimates $\mathcal{C}$, fused and deformed SDF $S$, and a sparse set of pixel-voxel correspondences, we want to label all 2D image pixels as $\{object,background\}$ and all voxels as $\{object,background,empty\}$. Note that even though the objects are placed on simple planar surfaces, achieving good quality segmentations is challenging because of one or more of the following reasons: (i) specularities; (ii) significant color variations on the object surface; (iii) errors in camera pose estimation; (iv) local errors in depth map estimates, (v) noisy PMVS point cloud; and (vi) error in support surface estimation. Another challenge is the computational complexity that arises as a result of dealing with pixels in all images and a dense voxel grid.

To keep computational complexity in check, we label superpixels instead of pixels. About $2000$ superpixels are computed for each image $I_i \in \mathcal{I}$ using SLIC \cite{achanta2012slic}. We denote the label assigned to the $i^{th}$ superpixel and voxel by $s_i$ and $v_i$, and the set of all pixels and voxels  by $\mathcal{S}$ and $\mathcal{V}$ respectively . We formulate our objective as a joint 2D-3D segmentation and minimize it using graph cuts with $\alpha$-expansion. The objective function is given by
\begin{multline}
E(s,v) = \sum_{i=1}^{|\mathcal{S}|}\psi_s(s_i) + \sum_{i=1}^{|\mathcal{V}|}\psi_v(v_i) + \sum_{(i,j)\in \mathcal{E}_{sv}}\psi_{sv}(s_i,s_j) \: + \\
\sum_{(i,j)\in \mathcal{E}_{ss}}\psi_{ss}(s_i,s_j) + \sum_{(i,j)\in \mathcal{E}_{vv}}\psi_{vv}(s_i,s_j)
\end{multline}
The superpixel unary term $\psi_s$ encodes object color, a scene prior that is informative about background superpixels, and consistency of image segmentations with volumetric reconstruction. The voxel unary term, $\psi_v$, encodes information about the empty regions using the SDF, location of background voxels using the scene prior pertaining to a flat support surface, and consistency of volumetric reconstruction with image segmentations. Super-pixel binary term, $\psi_{ss}$, encodes edge-aware smoothness between labeling of neighboring superpixels. Similarly, voxel binary term $\psi_{vv}$	 encodes smoothness in voxel labeling. Finally, superpixel - voxel pairwise term, $\psi_{sv}$, ensures consistency between surface voxels, and superpixels they project to in different views. Indirectly, $\psi_{sv}$ ensures consistent superpixel labeling across views. Next we describe the graph structure, initialization, and each of the unary and pairwise energy terms in detail.

\subsubsection{Graph Structure}

There is a node in the graph, $\mathcal{G}$, for every superpixel $\textbf{s}_i$ in $\mathcal{S}$ and every voxel $\textbf{v}_i$ in $\mathcal{V}$. We connect the neighboring voxels using a $6$-connected grid. We insert an edge between two superpixels if they both lie in the same image and share a boundary. We would ideally like to connect each superpixel with all voxels that project inside the superpixel, are in direct line of sight in the corresponding view, and lie on the surface of the object. However, this would require knowing the surface voxels and their visibility information. We can compute this information for a sparse set of points using semi-dense depth maps provided by LSD-SLAM. For each image we back-project the LSD-SLAM depth map and hash the points into the voxel grid $\mathcal{V}$. This gives us a sparse set of correspondences between superpixels and voxels. The region to be discretized by the voxel grid (our region of interest or ROI), is set to $median \pm 2$ standard deviations computed from the LSD-SLAM point cloud along each dimension. The resolution of the grid is chosen such that the largest dimension is divided into $200$ voxels.

\subsubsection{Initialization}
\begin{figure} [t]
\begin{center}
\includegraphics[width=0.9\linewidth]{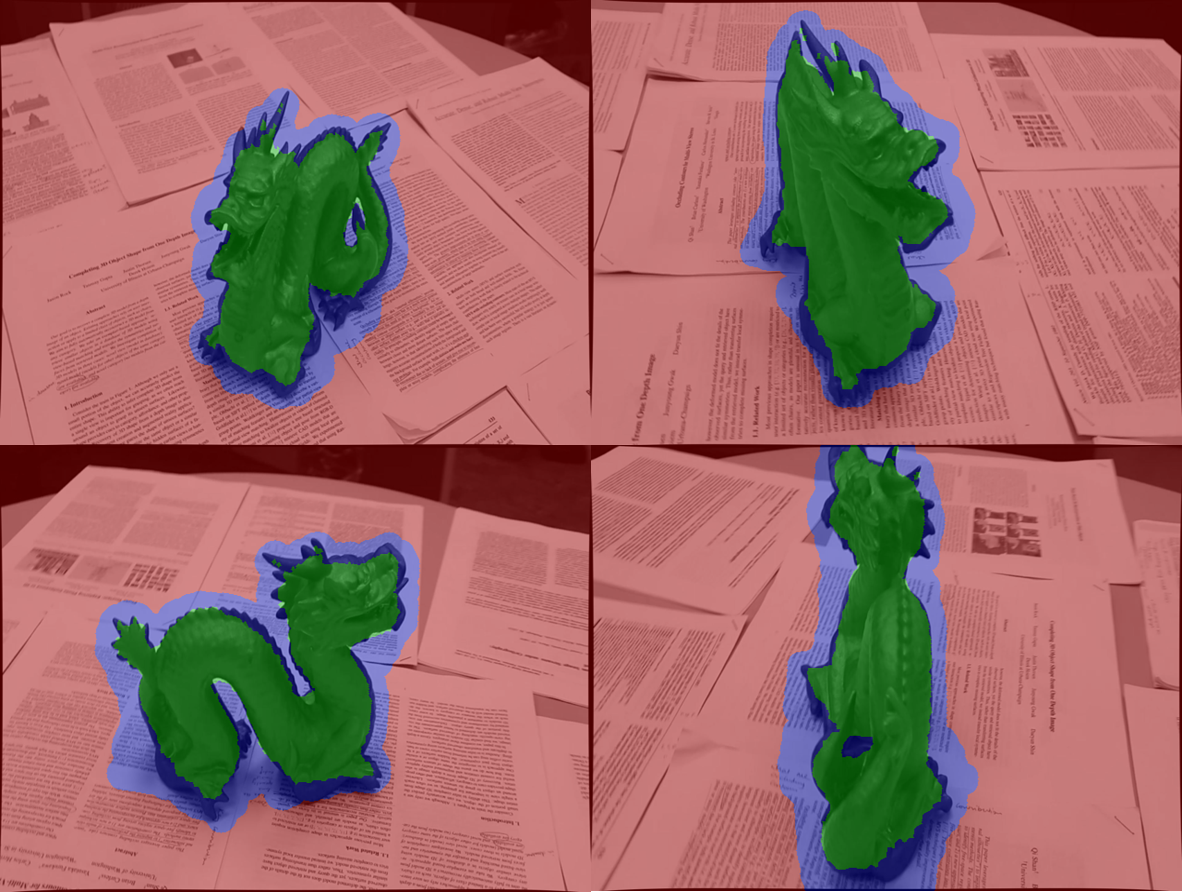}
\end{center}
\vspace{-0.1in}
   \caption{Pixel labels for 2D-3D Joint Segmentation are initialized using a trimap obtained from the deformed SDF. Only the labels of the blue colored pixels are optimized by graph cut.}
   \vspace{-0.1in}
\label{fig:trimapInit}
\end{figure}

We first extract the zero level set of the fused and deformed SDF and fit a plane to the points to identify the flat surface that supports the object. All voxels whose signed distance to the plane (negative denoting behind the plane) is less than a small positive theshold, $\tau$, are set to $background$. All voxels above the surface which contain at least one point from LSD-SLAM point cloud are marked as $object$ and the remaining ones are maked $empty$. We initialize the segmentation with a per-image trimap (see Fig.~\ref{fig:trimapInit}), computed by projecting an over-estimate, $\mathcal{U}$, and an under-estimate, $\mathcal{L}$, of the 3D volume on to each image. $\mathcal{U}$ is obtained by selecting the largest connected component from the set of voxels with $S(\textbf{v}) < 0.5$ and distance to the plane greater than $\tau$. We project $\mathcal{U}$ on to each image $I_i$ to get segmentation masks. Super-pixels that have more than $90\%$ pixels in the background region of the segmentation mask are permanently set to $background$. Next we create a set of voxels $\mathcal{L}=\{\textbf{v}\in \mathcal{U}|S(\textbf{v}) < -0.5\}$. $\mathcal{L}$ is projected on to each image plane to get segmentation masks and each superpixel with more than $90\%$ pixels in the foreground is permanently marked $object$. The remaining superpixels are randomly initialized. Note that only the superpixels that lie in this narrow band around the object silhouette are being solved for by graph cuts. A good initialization must include as few voxels on or below the plane in $\mathcal{U}$ as possible. For this $\tau$ is set to $0.05$ times the mean of the standard deviations along each dimension of the points used to fit the plane.

\subsubsection{Energy Terms}
\textbf{Super-Pixel Unary} $\psi_s$ comprises 3 components: $\psi_c$ encodes foreground and background colors; $\psi_{roi}$ ensures that superpixels which correspond to surfaces farther away from the ROI are assigned to background; and iii) $\psi_{sil}$ ensures consistency of silhouettes with volumetric reconstruction.

For the color term, separate GMMs are used to model the color distribution of foreground and background in Lab color space across all key-frame images. We learn both GMMs with 10 components and learn a full covariance matrix. $5000$ superpixels are randomly sampled across images from the currently labeled foreground or background regions and from them $50000$ pixels are sampled to learn each GMM model. Next, the posterior probabilities, denoted by $Prob(.)$, are computed for each pixel using the learned GMMs. In order to limit the influence of the color term, the probabilities are scaled and truncated by using the following function
\begin{equation}
M(x) = \min(5,-log(x))
\end{equation}
Let $\mathcal{P}(i)$ denote the set of pixels in the $i^{th}$ superpixel. Then the color based superpixel unary term is given by
\begin{equation}
\psi_c(s_i) = \frac{1}{|\mathcal{P}(i)|} \sum_{p \in \mathcal{P}(i)} M(\rm P_{gmm}(s_i|Color(p))
\end{equation}

To compute $\psi_{roi}$, we back-project pixels using the depth map and count how many lie inside the ROI. Let $f(i)$ denote the fraction of pixels in $\mathcal{P}(i)$ which fall outside the ROI. Then the ROI based term is given by
\begin{equation}
\psi_{roi}(s_i) =
\begin{cases}
f(i) & s_i=object \\
0      & s_i=background
\end{cases}
\end{equation}

For the silhouette based term, the current voxels labeled as object are used to render segmentation masks. Let $sil(i)$ denote the fraction of pixels in $\mathcal{P}(i)$ labeled as foreground. Then silhouette consistency term is given by
\begin{equation}
\psi_{sil}(s_i) =
\begin{cases}
1-sil(i) & s_i=object \\
sil(i)      & s_i=background
\end{cases}
\end{equation}

Finally $\psi_s(s_i) = \psi_c(s_i) + \lambda_{roi}\psi_{roi}(s_i) + \lambda_{sil}\psi_{sil}(s_i)$. For our experiments we used $\lambda_{roi} = 10$ and $\lambda_{sil} = 1$. In our experiments, we found that using an average of $\psi_{sil}$ over all past iterations of graph cut leads to more stable solutions.

\begin{figure*}[t]
\begin{center}
\includegraphics[width=1\linewidth]{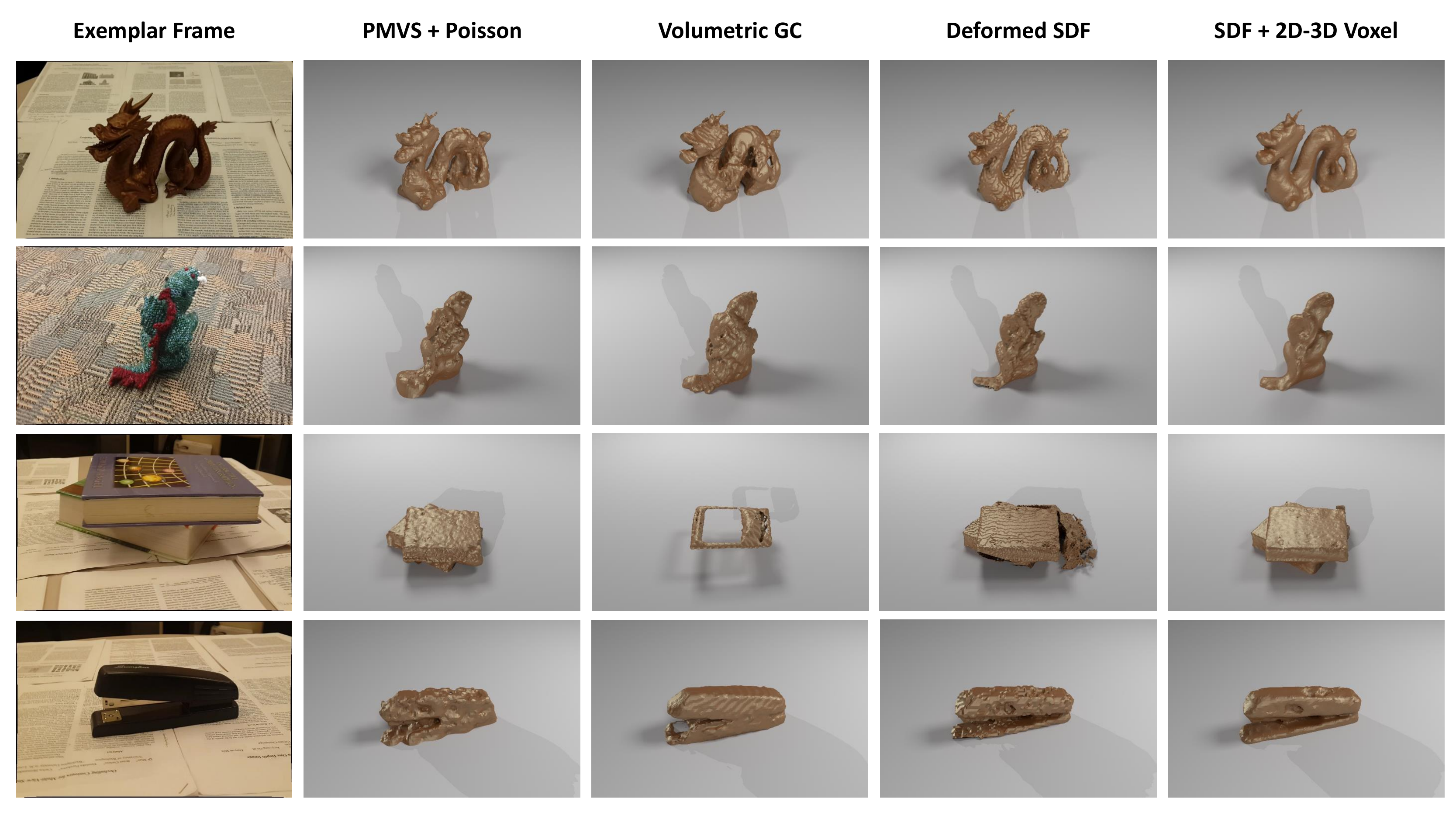}
\end{center}
\vspace{-0.1in}
   \caption{Comparison of reconstructions produced by our system ``SDF + 2D-3D Voxel'' with other baseline methods on videos in Object-Videos dataset. Our approach incorporates: dense depth maps with an improved cost function and improved signed distance function for fusion; sparse patch-based stereo 3D points (PMVS); and joint 2D-3D segmentation based on color, edge, and depth cues.  By using a broad set of techniques, our system avoids many of the errors and artifacts inherent in any individual. Note that our method produces smoother and more detailed reconstructions as compared to the baseline methods while being more robust to errors in PMVS point cloud and depth estimation errors that show their effect in ``PMVS + Poisson'' and ``Deformed SDF''. Also unlike ``Volumetric GC'', our approach is less sensitive to the color model, as evident from the \textit{books} example.}
   \vspace{-0.1in}
\label{fig:baselineCompare}
\end{figure*}

\textbf{Voxel Unary} $\psi_v$ has 3 components: $\psi_{sdf}$ encodes the information contained in the SDF, $S$; $\psi_{scene}$ encodes our scene prior; and $\psi_{carve}$ ensure that voxel labeling is consistent with superpixel labeling. Note that SDF helps in identifying empty regions ($S(\textbf{v})>0$) but cannot distinguish between object of interest and other background regions, such as those below the support surface or regions further away from the object, since SDF attains a positive value in both cases. This complementary information is provided by the other two components, $\psi_{scene}$ and $\psi_{carve}$.

First, for each voxel $\textbf{v}$, a normalized distance to the fitted plane is computed as follows
\begin{equation}
d_{n}(\textbf{v}) = \frac{1}{1+e^{-d(\textbf{v})/\sigma_{d}}}
\end{equation}
where $d(\textbf{v})$ denotes the distance of $\textbf{v}$ from the plane and $\sigma_d$ is the standard deviation of the distance of all voxels from the plane. Next, all voxels with $d_n(\textbf{v}) < 0.3$ are permanently set to $background$. All voxels with $S(\textbf{v}) > 0.9$ are permanently set to $empty$. For the remaining voxels, $\textbf{v}$ the $SDF$ based energy term is given by
\begin{equation}
\psi_{sdf}(v_i) =
\begin{cases}
 S(\textbf{v}) & v_i=object \\
 S(\textbf{v}) & v_i=background \\
-S(\textbf{v}) & v_i=empty
\end{cases}
\end{equation}

The scene prior energy term, $\psi_{scene}$, penalizes labeling of voxels above the support surface as background and those below the surface as either $object$ or $empty$. It is defined as
\begin{equation}
\psi_{scene}(v_i) =
\begin{cases}
\mathbbm{1}(d_{n}(\textbf{v})<0.5) \; (1-d_{n}(\textbf{v}))   & v_i=object \\
\mathbbm{1}(d_{n}(\textbf{v})>0.5) \; d_{n}(\textbf{v})   & v_i=background \\
\mathbbm{1}(d_{n}(\textbf{v})<0.5) \; (1-d_{n}(\textbf{v}))  & v_i=empty
\end{cases}
\end{equation}

In order to compute $\psi_{carve}$, we first perform silhouette carving to get a set of voxel $\mathcal{V}_{carve}$ which project on the superpixels labeled $object$ in all key-frame views. Let the complementary set be $\mathcal{V}_{carve}^c = \mathcal{V}\setminus\mathcal{V}_{carve}$. $\psi_{carve}$ is then given by
\begin{equation}
\psi_{carve}(v_i) = \mathbbm{1}(\textbf{v}_i\in \mathcal{V}_{carve}^c)\mathbbm{1}(v_i=object)
\end{equation}

The final voxel unary for the variable voxels is given by $\psi_v(v_i)~=~\lambda_{sdf}\psi_{sdf}(v_i) + \lambda_{scene}\psi_{scene}(v_i) + \lambda_{carve}\psi_{carve}(v_i)$. For all our experiments we have set $\lambda_{sdf}=4$, $\lambda_{scene}=1$, and $\lambda_{carve}~=~1$. Similar to $\psi_{sil}$, we also use the average $\psi_{carve}$ over all past iterations.

\textbf{Super-pixel Binary} term $\psi_{ss}$ imposes edge-aware smoothness constraints on the superpixel labeling. For a pair of superpixels with indices $(i,j)$ that share boundary, with the set of boundary pixels denoted by $\mathcal{B}(i,j)$, the superpixel pairwise energy term is given by
\begin{equation}
\psi_{ss}(s_i,s_j) = \mathbbm{1}(s_i\neq s_j)\exp\left(-2\sum_{\textbf{p}\in \mathcal{B}(i,j)}Edge(\textbf{p})\right)
\end{equation}
where $Edge(\textbf{p})$ denotes the contour edge strength of pixel $\textbf{p}$ obtained using Structured Edge Detector \cite{dollar2013structured}.

\textbf{Voxel Binary} term $\psi_{vv}$ imposes smoothness constraints on the voxels. For a pair of voxels indexed by $(i,j)\in \mathcal{E}_{vv}$, the voxel binary term is defined as $\psi_{vv}(v_i,v_j) = \mathbbm{1}(v_i\neq v_j)$.

\textbf{Super-pixel - Voxel Pairwise} term $\psi_{sv}$ ensures that for every superpixel and voxel with an edge between them, with indices $(i,j)\in \mathcal{E}_{sv}$, superpixel and voxel labels are consistent with each other. Note that $\mathcal{E}_{sv}$ only connects surface voxels with superpixels. The surface voxels can never be $empty$ hence both $s_i$ and $v_j$ lie in $\{object,background\}$. Given this restriction on the labels $\psi_{sv}(s_i,v_j) = \mathbbm{1}(v_i\neq v_j)$.

\subsubsection{Details of $\alpha$-expansion}
We run $5$ iterations of optimization with $3$ $\alpha$-expansion steps per iteration or until the sum of the number of label swaps for all expansion moves  in an iteration falls below $6000$, whichever comes first. To make the set of labels the same for pixels and voxels, we assign a very high cost to $empty$ for any superpixel. The color models are updated in each iteration using the current superpixel labeling hence $\psi_c$ needs to be recomputed in each iteration. Besides this only $\psi_{sil}$ and $\psi_{carve}$ are updated in each iteration since they depend on the current voxel map and segmentation masks.

\subsection{Post-processing} \label{sec:postProcess}
As an important post-processing step, we select the largest connected component from the voxels labeled as $object$. While this could be used to generate a surface mesh of the object directly, we found that qualitatively better results are obtained by using it as a mask to set $S(\textbf{v})$ to $1$ for all $\textbf{v}$ such that $v\neq object$, extract its zero-level set, and perform mesh smoothing using \cite{zhang2006vertex}.

\section{Experiments}

\begin{figure*} [t]
\begin{center}
\includegraphics[width=0.9\linewidth]{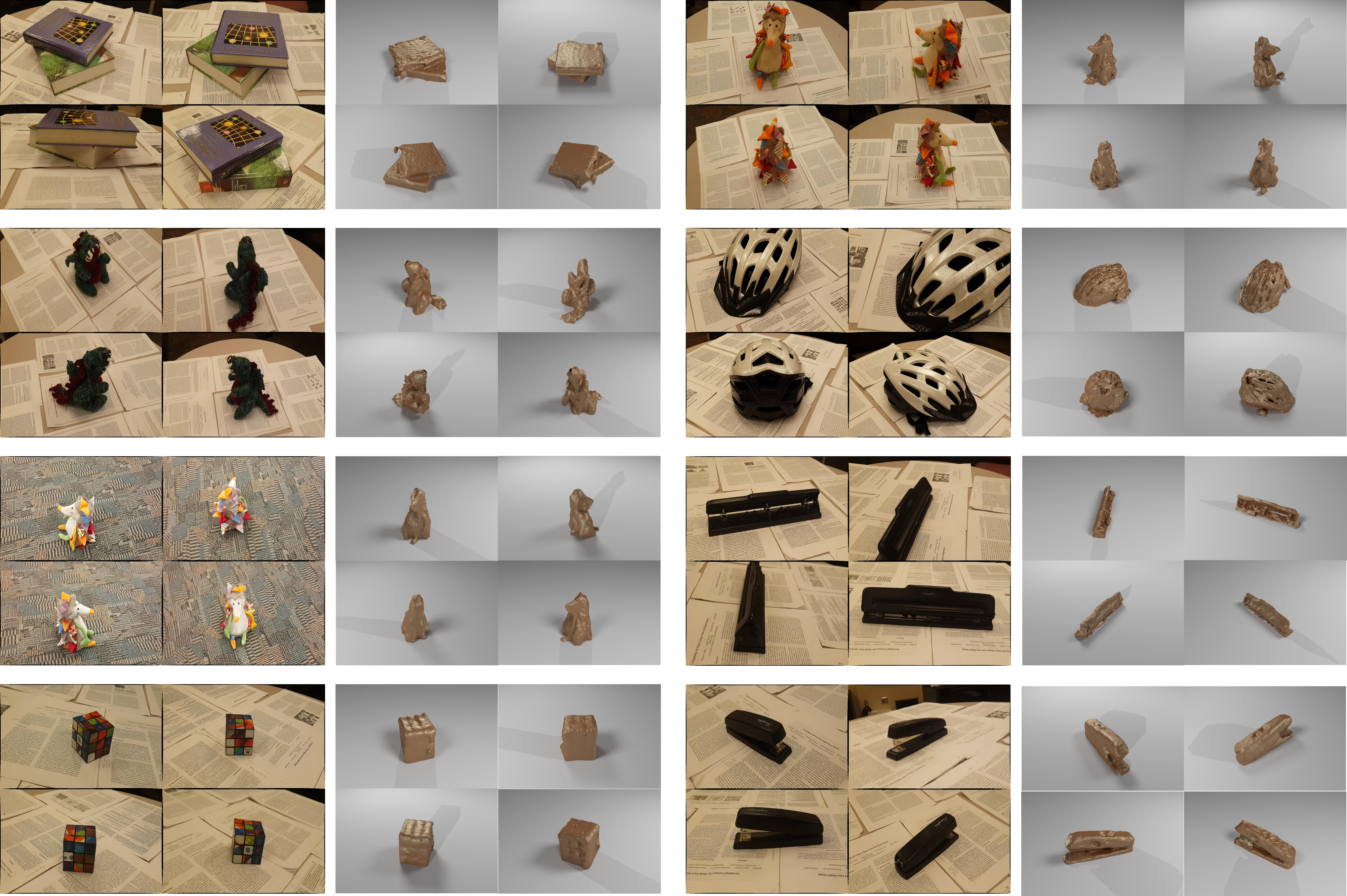}
\end{center}
	\vspace{-0.1in}
   \caption{Qualitative reconstruction results for more videos in Object-Videos dataset. Our approach produces naturally smooth surface reconstruction of objects with varying complexity of shapes and material properties.}
   \vspace{-0.1in}
\label{fig:qualWall}
\end{figure*}

We evaluate our system qualitatively (Fig.~\ref{fig:qualWall} and Fig.~\ref{fig:baselineCompare}) with displays of 3D reconstructed objects from our Object-Videos dataset.  We also quantitatively evaluate dense depth map estimation (Fig.~\ref{fig:plot}) with the RGBD Scenes V2 Dataset~\cite{henry2013patch} and segmentation accuracy (Tab.~\ref{tab:fgIOU}) on key frames from the Object-Videos dataset.


\subsection{Object Reconstruction}

Our Object-Videos dataset consists of 12 videos of 10 objects captured using a commercial mobile phone camera Samsung Galaxy S4.  Many of the objects have complex shapes, low-texture surfaces, and specular materials.  Thus, while many graphics and vision papers use carefully designed experimental setups and/or objects with smooth Lambertian surfaces that satisfy model assumptions, we attempt to reconstruct common objects filmed with a typical camera in a casual process.  In Fig.~\ref{fig:baselineCompare}, we show results of our method ``SDF + 2D-3D Voxel'' with comparisons to baseline methods: (i) Poisson surface reconstruction~\cite{kazhdan2006poisson} using PMVS~\cite{furukawa2010accurate} point cloud; (ii) Volumetric graph cut method of ~\cite{vogiatzis2005multi}; and (iii) Zero level set of deformed SDF $S$ after selecting appropriate region using aggressive and conservative thresholds on $S$.

\subsection{Dense Depth Estimation}

\begin{figure} [t]
\begin{center}
\includegraphics[width=1\linewidth]{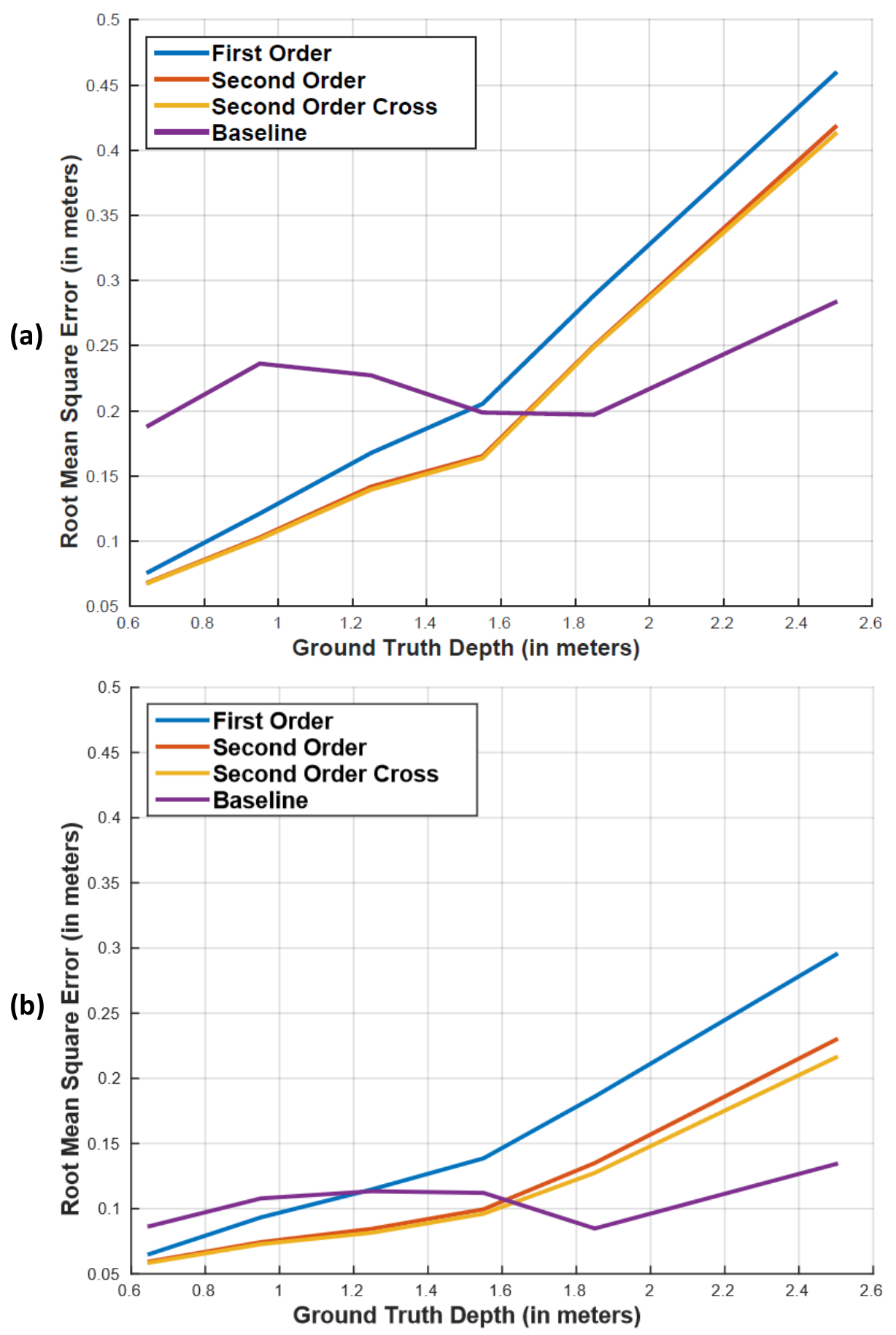}
\end{center}
	\vspace{-0.1in}
   \caption{Evaluation of dense depth map computation on RGBD Scenes v2 dataset with - (a) 100, and (b) 1000 measurements. Our algorithm is better at reconstructing nearby depths because of nonlinear discretization used for cost volume construction which leads to finer resolution at smaller depths. The object of interest is typically close to the camera.}
   \vspace{-0.1in}
\label{fig:plot}
\end{figure}

To evaluate our dense depth map estimation procedure we use the available camera poses to estimate depth maps for five selected key frames of each video in the RGBD Scenes V2 dataset. We compare three different smoothness priors: first order ($\|\nabla\xi(u)\|_2^2$); second order (eq.~\ref{eq:bendingEnergy} without the cross term); and second order rotation invariant bending energy (eq.~\ref{eq:bendingEnergy}). We also evaluate the effect of increasing the number of measurements (known depth points) from 100 to 1000 (see Fig.~\ref{fig:rgbd_qual} for a qualitative comparison on RGBD Scenes V2 dataset). We compare to bilinear interpolation of known depth values as a baseline. For indoor scenes that consist of many planar surfaces, the baseline works quite well, but it provides poor estimates for objects that have curved surfaces. Fig.~\ref{fig:depthNormal} shows the qualitative comparison of depth estimation with different smoothness priors on 2 videos from our Object-Videos dataset. 

In Fig.~\ref{fig:plot}, we plot root mean squared error as a function of ground truth depth. All the variants of our approach beat the baseline for small depths, which is most relevant for our application. For large depths, non-linear discretization used for the cost volume results in high quantization errors, causing the dense depth estimation to underperform the baseline. While inverse depth and log space discretization are common, in our experiments we found that scaling  $\{(0.2 + 0.8\times i/l)^5\}_{i=1}^l$ to the expected range of depth values, where $l$ is the number of discrete values, performed best. 

Among smoothness priors, the rotation invariant second order bending energy performs the best, beating the second order bending energy by a small but noticeable amount. However, the latter may be preferred because it requires less computation.  Both the second order energies perform significantly better than the first order energy.

Finally, as the number of measurements increase the error reduces, as expected, especially for regions further from the camera.

\subsection{Object Segmentation}

\begin{table*}[t]
  \centering
  \resizebox{\textwidth}{!}{%
  \begin{tabular}{|c|cccccccccccc|c|c|}
  \hline
                              & \textbf{dragon} & \textbf{stapler} & \textbf{rubik\_cube} & \textbf{paper\_punch} & \textbf{keyboard} & \textbf{teabox} & \textbf{books} & \textbf{helmet} & \textbf{hedgehog1} & \textbf{hedgehog2} & \textbf{godzilla1} & \textbf{godzilla2} & \textbf{mean} & \textbf{median}              \\ \hline \hline
  \textbf{Fused SDF}          & 0.86         & 0.79          & 0.83              & 0.73               & 0.74           & 0.74         & 0.88        & 0.84          & 0.82            & 0.73            & 0.82            & 0.78            & 0.80       & \multicolumn{1}{c|}{0.80} \\
  \textbf{Deformed SDF}       & 0.87          & 0.82          & 0.85              & 0.81                & 0.76           & 0.54         & 0.81        & 0.79         & 0.63            & 0.80            & 0.58            & 0.82            & 0.76       & \multicolumn{1}{c|}{0.81}  \\
\hline \hline
  \textbf{Volumetric GC}      & 0.90		& 0.86	 	& 0.76		& 0.88		& 0.68		& 0.85		& 0.35		& 0.37		&	0.86		& 0.86		& 0.81		& 0.90		& 0.76 		& \multicolumn{1}{c|}{0.85}  \\		
  \textbf{2D GC}              & 0.90         & 0.90          & 0.95              & 0.93               & 0.76           & 0.94         & 0.88        & 0.84         & 0.83            & 0.88            & 0.80            & 0.91            & 0.88       & \multicolumn{1}{c|}{0.89}  \\
  \textbf{2D-3D GC Pixel:}           & 0.91         & 0.90          & 0.95              & 0.93               & 0.78            & 0.87         & 0.92        & 0.87         & 0.85            & 0.85            & 0.83            & 0.91            & 0.88       & \multicolumn{1}{c|}{0.88} \\
  \textbf{2D-3D GC Voxel}    & 0.91         & 0.90          & 0.93              & 0.88               & 0.73           & 0.84         & 0.92        & 0.88         & 0.87            & 0.82            & 0.80            & 0.88            & 0.86        & \multicolumn{1}{c|}{0.88} \\
  \textbf{GC Select}          & 0.90         & 0.89          & 0.91              & 0.88               & 0.74           & 0.82         & 0.92        & 0.87         & 0.86            & 0.80            & 0.79            & 0.86            & 0.85       & \multicolumn{1}{c|}{0.87} \\
\hline \hline
  \textbf{SDF + 2D-3D Voxel} & 0.90         & 0.89          & 0.91              & 0.88               & 0.73           & 0.82         & 0.91        & 0.87         & 0.86            & 0.79            & 0.79            & 0.85            & 0.85       & \multicolumn{1}{c|}{0.87} \\ \hline
  \end{tabular}}
  \vspace{-0.1in}
  \caption{Foreground IoU}
  \vspace{-0.1in}
   \label{tab:fgIOU}
\end{table*}

\begin{figure*} [t!]
\begin{center}
\includegraphics[width=0.9\linewidth]{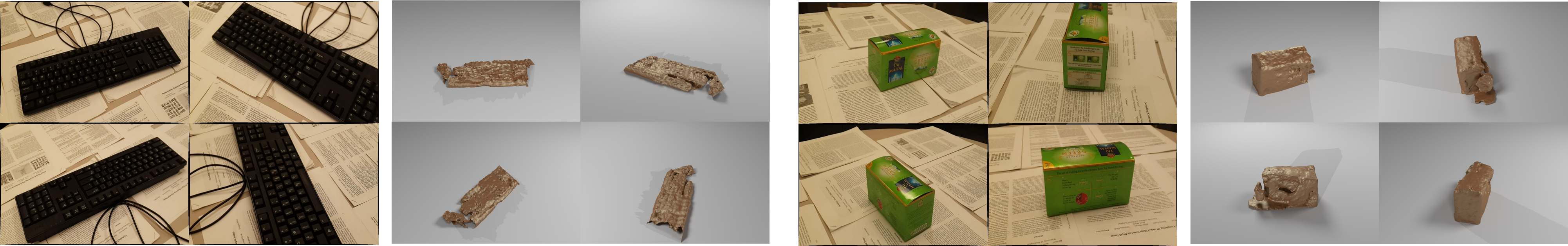}
\end{center}
	\vspace{-0.1in}
   \caption{Failure cases which are mainly caused due to camera pose estimation errors. Pose errors affect all stages in the pipeline. Incorrect support surface estimation is an issue for flat objects since there is little evidence to distinguish between object and background regions.}
   \vspace{-0.1in}
\label{fig:failCases}
\end{figure*}

We also compare variants of our algorithm on 2D pixel segmentation.  These results give a sense of 3D segmentation and reconstruction quality, but multiple 3D segmentations are consistent with a single 2D segmentation (e.g., the visual hull and the true shape have the same silhouettes) and different kinds of 2D segmentation errors have different impact on 3D reconstruction. In Table~\ref{tab:fgIOU}, we compare segmentation performance, measured by intersection over union (IoU) of ground truth and estimated segmentation masks over annotated keyframes, for several variants of our technique:
\begin{itemize}
\item[] \textbf{Fused SDF: } Segmentations obtained by backprojecting the largest connected component of interior voxels above the fitted plane from the signed distance function (SDF) created from depth maps.
\item[] \textbf{Deformed SDF: } Same as above, but after deforming the SDF so that its surface lies close to PMVS points.
\item [] \textbf{Volumetric GC: } Our implementation of volumetric graph cut method of \cite{campbell2010automatic}.
\item[] \textbf{2D GC: } Image graph cuts co-segmentation using color terms computed from multiple images but without voxel segmentation, the pixel-voxel constraints, or the $\psi_{sil}$ term.
\item[] \textbf{2D-3D GC Pixel: } Pixel segmentations resulting from our joint 2D-3D graph cuts method.
\item[] \textbf{2D-3D GC Voxel: } Backprojects segmented voxels using our joint 2D-3D method.
\item[] \textbf{SDF + 2D-3D Voxel: } Backprojects the 3D volume obtained by slightly dilating the voxels obtained from 2D-3D segmentation, intersecting with Deformed SDF, and smoothing.  This method provides the best qualitative results and is used for reconstruction results.
\end{itemize}

The SDF deformation yields better results in most cases, but for \textit{teabox}, \textit{books}, \textit{helmet}, \textit{hedgehog1}, and \textit{godzilla1}, the results are much worse due to large errors in camera pose estimation that decreased accuracy of the PMVS point cloud used for deformation.  The PMVS results are included in the supplementary video.  Our graph cuts methods are robust to these errors, outperforming ``Fused SDF'' and ``Deformed SDF'' by large margins.

Our joint 2D-3D segmentation (``2D-3D GC Pixel'') performs equally or better than 2D-only segmentation (``2D GC'') in all but two cases, supporting the value of joint segmentation.  The main case in which 2D-only outperforms is \textit{teabox} in which  errors in camera pose estimation harm the 2D-3D result.  Although backprojecting voxels from the 2D-3D segmentation (``2D-3D GC Voxel'') slightly underperforms the 2D-3D pixel segmentation (partly due to coarser voxel discretization), the voxel-based 3D model is better than that obtained by shape carving from 2D segmentations because shape carving is sensitive to errors in individual images.  Our most qualitatively pleasing 3D models are produced by combining meshes from ``Deformed SDF'' and ``2D-3D GC Voxel'', but when backprojected to images, the resulting silhouettes are slightly less accurate than ``2D-3D GC Voxel''.

As an alternative to graph cut optimization, we tried Spectral Matting \cite{levin2008spectral} for performing independent image segmentations using the trimap. Spectral Matting first identifies a basis set of matting components where each component is obtained as a linear combination of a Laplacian matrix. Then it uses the trimap initialization to assign each component to foreground or background and constructs an $\alpha$-matte. However, we found that in order to obtain consistent segmentations across views it is necessary to compute eigenvectors of a prohibitively large Laplacian matrix defined over pixels in all key-frames and all voxels.



\section{Failure Modes and Limitations}
Based on qualitative (see Fig.~\ref{fig:failCases}) and quantitative evaluation, we have identified a set of failure modes for our approach -

A major source of error that affects all stages of our system is camera pose estimation. Camera pose estimates directly affect the accuracy of constructed \textit{cost volume} and depth measurements for computing dense depth maps, accuracy of PMVS point cloud, alignment of depth maps during volumetric fusion, deformation of SDF using PMVS point cloud, and superpixel-voxel consistency constraints during joint 2D-3D segmentation. The pose estimation errors are largely due to severe occlusion and breaking of brightness constancy assumption due to textureless and specular surfaces. The videos in Object-Videos dataset are also collected in a casual fashion with blurry frames and large displacement between consecutive frames.

Secondly, errors in estimated depth maps due to specular and textureless surfaces adversely affect quality of the fused SDF which is the driving force behind the 2D-3D segmentation mechanism. We have demonstrated some degree of robustness in reconstructing textureless and specular surfaces such as in \textit{stapler}, \textit{paper-punch}, \textit{teabox}, and \textit{helmet}, but explicit removal of specularities would further improve performance.

As a byproduct of the above two, incorrect estimation of the support surface is a main source of error while reconstructing flat objects like the \textit{keyboard}.

Our approach also has some limitations. Our method was targeted towards reconstruction of small objects for 3D printing or augmented reality applications and hence applies to a scale of objects and scenes which can afford computation with a discrete grid of voxels. We are also limited by computation of the cost volume for depth estimation which grows linearly in the number of discrete depth values used. We limit our selves to planar support surfaces with limited background clutter. Finally, our method cannot recover fine geometric details such as the scale pattern on the \textit{dragon}. Such detail recovery would require shape from shading and material analysis which are still open research problems. 

\section{Conclusion}

In this work, we proposed a system for 3D reconstruction of an object from a video taken with hand-held mobile phone camera.  Our three major contributions are: (1) improved objective function for dense depth map computation; 2) robust estimation of an implicit surface using a softmax signed distance function with zero-crossing correction; and (3) a method for joint $2D$-$3D$ segmentation. Through qualitative and quantitative results we demonstrate robustness to textureless surfaces, specularities, and errors in camera pose estimation. 
Potential directions for future work include extending the proposed approach for category specific reconstruction using data driven priors and recovering high frequency details in the reconstruction through shape-from-shading and material analysis.

\section*{Acknowledgements}
This research was supported in part by NSF Award 14-21521. We are also thankful to David Forsyth for helpful discussion on linearized bending energy and smoothness priors, and to Jason Rock for suggesting the region of interest based component of superpixel unary.

\bibliographystyle{acmsiggraph}
\nocite{*}
\bibliography{ShapeFromVideo}
\end{document}